\documentclass[sigconf]{acmart}
\usepackage{adjustbox}
\AtBeginDocument{%
  }

\setcopyright{acmlicensed}
\copyrightyear{2018}
\acmYear{2018}
\acmDOI{XXXXXXX.XXXXXXX}
\acmConference[CIKM '26]{35th International ACM Conference on Knowledge and Information Management (Under Review)}{June 03--05,
  2018}{Woodstock, NY}
\acmISBN{978-1-4503-XXXX-X/2018/06}

\usepackage{baostyle}
\usepackage{graphicx}
\usepackage{enumitem}

\begin{document}
\title{Didact: A Cross-Domain Capability Discovery System for Defence}

\author{Aarya Bodhankar}
\affiliation{%
  \institution{University of New South Wales, Sydney, Australia}
  \country{Australia}}
\email{a.bodhankar@unsw.edu.au}

\author{Aditya Joshi}
\affiliation{%
  \institution{University of New South Wales, Sydney, Australia}
  \country{Australia}}
\email{aditya.joshi@unsw.edu.au}

\author{Bao Gia Doan}
\affiliation{%
  \institution{University of New South Wales, Sydney, Australia}
  \country{Australia}}
  \email{bao.doan1@unsw.edu.au}

\author{Thomas Marchant}
\affiliation{%
  \institution{Cyndr.ai, Australia}
  \country{Australia}}
  \email{tom@cyndr.ai}

\author{Oscar Leslie}
\affiliation{%
  \institution{Cyndr.ai, Australia}
  \country{Australia}}
\email{oscar@cyndr.ai}
\author{Flora Salim}
\affiliation{%
  \institution{University of New South Wales, Sydney, Australia}
  \country{Australia}}
\email{flora.salim@unsw.edu.au}
% \author{Aditya Joshi}
% \affiliation{%
%   \institution{University of New South Wales}
%   \city{Sydney}
%   \state{NSW}
%   \country{Australia}}
% \email{aditya.joshi@unsw.edu.au}

% \author{Bao Gia Doan}
% \affiliation{%
%   \institution{University of New South Wales}
%   \city{Sydney}
%   \state{NSW}
%   \country{Australia}}
% \email{bao.doan1@unsw.edu.au}

% \author{Thomas Marchant}
% \affiliation{%
%   \institution{Cyndr AI}
%   \country{Australia}}
% \email{tom@cyndr.ai}

% \author{Oscar Leslie}
% \affiliation{%
%   \institution{Cyndr AI}
%   \country{Australia}}
% \email{oscar@cyndr.ai}

% \author{Flora Salim}
% \affiliation{%
%   \institution{University of New South Wales}
%   \city{Sydney}
%   \state{NSW}
%   \country{Australia}}
% \email{flora.salim@unsw.edu.au}
\renewcommand{\shortauthors}{Bodhankar et al.}

\begin{abstract}
Policymakers in defence and defence-aligned sectors must monitor rapidly evolving research alongside sector priorities relevant to operational and strategic needs. In practice, these sources are fragmented across heterogeneous formats, disjoint repositories, and siloed update streams, making capability discovery slow and difficult to audit. We present Didact, a prototype that integrates publicly available defence reports and policy documents from Australia with a purpose-built knowledge graph derived from Australian research publications. Didact provides natural language conversations for policy-oriented workflows, and leverages a composite retrieval-augmented generation (RAG) pipeline. A key feature of Didact is an interactive Evidence Rail that visualises retrieved evidence and source relationships. Our evaluation of the output quality and runtime of Didact highlights its utility. While Didact has been co-developed as an academia-industry project for the Australian context, it is adaptable to other domains where knowledge is similarly fragmented. A demonstration video is available here: [\textcolor{blue}{\underline{\hyperlink{https://youtu.be/KRtlq3LQ9WA}{video}}}].
\end{abstract}



\maketitle

\section{Introduction}
Research outputs, policy documents, and defence reports tend to live in separate silos. When government officials and defence specialists search for a specific technology, capability, or emerging research theme, they must manually reconcile academic publications, policy context, and operationally relevant reports. This fragmentation limits situational awareness---analysts can locate individual artefacts, but it is much harder to see the broader landscape of priorities, organisations, technologies. Retrieval-augmented generation (RAG) with large language models (LLMs) has become a widely adopted approach for building domain-grounded conversational systems, as retrieved context can both constrain generation and surface supporting sources \cite{lewis2020retrieval,gao2023retrieval}. However, most RAG systems are built around a single document collection and a narrow class of fact-seeking queries. Capability discovery in the defence domain demands synthesis across heterogeneous sources, tighter control over which sources are retrieved, and interfaces that let users examine evidence rather than simply accept a generated answer.

Didact is the first-of-its-kind tool that demonstrates a multi-source composite RAG system which allows for cross-domain capability discovery. The two domains, Defence and academic research, are collectively leveraged using unstructured document retrieval over curated defence material and graph retrieval over Australian research metadata. In particular, a RAG-based conversational LLM agent orchestrates these retrieval tools, allowing users to ask questions (such as which Australian researchers are active in a given capability area, which documents support a particular strategic claim, or how research activity connects to policy priorities). Given the Defence context, a core design principle is evidence-grounded synthesis under source-level control. Didact organises defence documents in two access levels (on similar lines as security clearance levels in Australia~\footnote{\url{https://www.agsva.gov.au/about/security-clearance-definitions}}), keeps graph and document retrieval as distinct tools, and attaches an evidence rail to each response. The evidence rail comprises the retrieved documents, graph entities, and relationships that informed an answer, enabling users to trace from a generated summary directly to the underlying evidence. The graph entities are themselves interactive and all the underlying choices (model provider, etc.) are configurable in an admin dashboard. Didact represents an innovative prototype of artificial intelligence being deployed for cross-domain capability discovery using natural language conversations. It contextualises typical RAG-based systems for Defence in three key ways: (1) It identifies the need for multi-domain coverage; (2) It maintains source separation; and (3) It provides a detailed evidence-centred user interface for inspecting cited documents and graph relationships. In other words, our tool is knowledgeable (\textit{i.e.}, `didact') about the requirements of a RAG-based conversation agent in the Defence setting, lending Didact its name.

\begin{figure*}[t]
    \centering
    \includegraphics[width=0.8\linewidth]{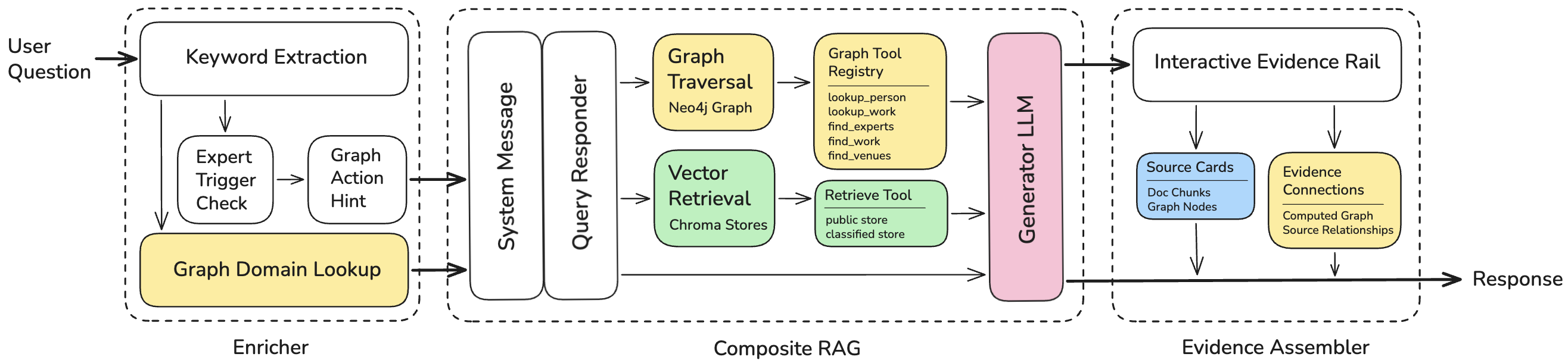}
  %  \fbox{\parbox{0.92\linewidth}{\centering \includegraphics[\linewidth]{didact-demo-1.png}}}
    \caption{Architecture of Didact. Didact combines tiered document retrieval with structured graph retrieval. The LLM agent calls retrieval tools, while the interface exposes generated answers together with cited evidence and relationships.}
   % \Description{Placeholder for the Didact system architecture figure.}
    \label{fig:architecture}
\end{figure*}

\section{Related Work}
RAG improves LLM factuality by retrieving relevant context at inference time and giving users a path from generated text back to supporting sources \cite{lewis2020retrieval,gao2023retrieval}. This is particularly important in high-stakes domains where unsupported synthesis can erode trust. Work on graph-enhanced RAG augments document retrieval with structured knowledge graphs to enable multi-hop reasoning, richer context integration, and more interpretable retrieval pathways \cite{peng2025graph,zhang2025survey}. Security-focused systems have likewise used knowledge graphs to improve traceability and interpretability over heterogeneous evidence sources \cite{sharma2025enhancing}. Didact builds on these directions but targets a different operational setting. Rather than treating retrieval as a document-ranking problem, it separates document and graph retrieval into distinct agent-callable tools, preserves access-control boundaries between document stores, and surfaces evidence as an interactive part of the interface. This focus on cross-domain capability discovery positions the system closer to an analyst workspace than a conventional question-answering assistant.

\section{Architecture}
As illustrated in Figure~\ref{fig:architecture}, Didact is implemented as a pipeline that takes a user question as input and generates a response. The response consists of a textual response along with an evidence rail consisting of document chunks and relevant subgraphs. Prior turns are replayed into the workflow to enable multi-turn conversation, while responses are streamed to the client via SSE. 
\subsection{Data Sources}
The knowledge base spans two source types: curated Defence documents and a research knowledge graph, which support complementary retrieval modalities. The Defence document collection comprises publicly available Australian Defence reports and policy releases, curated by defence domain experts. For this demonstration, the documents are randomly partitioned into two access levels `public' and `classified' to exercise the clearance model typical in Defence applications, and demonstrate source isolation. Source isolation is maintained through separate Chroma~\cite{chroma2023} vector stores, one per access level. The documents themselves are ingested through a document-processing pipeline, split into 1500-character chunks with 20-character overlap, and assigned UUIDs at both document and chunk level for traceability. The research knowledge graph is purpose-built from OpenAlex~\cite{priem2022openalex} metadata. The graph represents publications as Work nodes linked to authors, organisations, venues, domain tags, and countries through typed relationships that capture authorship, affiliation, venue assignment, topical annotation, and geographic location. Source data is filtered for Australian authorship, enriched with institution country codes from the Research Organisation Registry (ROR)~\footnote{\url{https://ror.org/}}, and annotated with domain taxonomy tags derived from the partner ontology. The academic knowledge graph is open metadata and is available at both levels. 

\subsection{Enricher}
Didact performs a pass of the Enricher against the knowledge graph. It extracts domain-relevant keywords from the user question by performing graph domain lookup, matches them against taxonomy tags. For researcher-discovery queries, this stage can also prime the agent toward graph retrieval before generation begins.  In particular, the Enricher extracts domain keywords from the query and matches them against \texttt{DomainTag} nodes in Neo4j~\cite{neo4j}. When matches are found, a domain context block is prepended to the system prompt. For researcher-discovery queries, an additional graph-action instruction can be injected, pre-resolving the relevant graph operation before the agent begins reasoning. Enricher injects the relevant context into a system message.
\subsection{Composite RAG}
Composite RAG is implemented using LangGraph's Orchestrator, which delegates to a graph-structured agent engine~\cite{langgraph2024} to assemble and run the conversational workflow. Unstructured document content is retrieved through dense vector similarity search \cite{Karpukhin_2020}, while the knowledge graph is queried through structured operations for entity and relationship reasoning \cite{Hogan_2021}. This separation allows Didact to answer both evidence-seeking questions over documents and discovery-oriented questions about researchers, organisations, venues, and topics. The graph representation of retrieved sources is computed once per response and exposed to the user as an interactive evidence graph.  Composite RAG is built around three conceptual nodes. \textit{QueryResponder} determines whether the user query requires tool calls or can be answered directly. \textit{ToolNode} executes the selected retrieval tools. \textit{ResponseGenerator} receives the tool results and produces a final answer with citations. \textit{ToolNode} exposes two retrieval tools: vector retrieval, which returns ranked chunks from the relevant Chroma store; and graph retrieval, which supports five structured operations---finding works by topic, finding experts by topic, finding venues by topic, looking up an author's publications by name, and retrieving a specific work by identifier.

The Generator LLM is selected by administrators who specify LLM provider, model, and system prompt. In general, Composite RAG in Didact is highly adaptable in that the LLM can run as a self-hosted container or delegate to external provider APIs, while document and graph retrieval components remain modular so that new domains can be added with minimal configuration.

%\textit{GraphEnricher} runs before the workflow is invoked.

\begin{figure*}[t]
    \centering
    \includegraphics[width=0.4\linewidth]{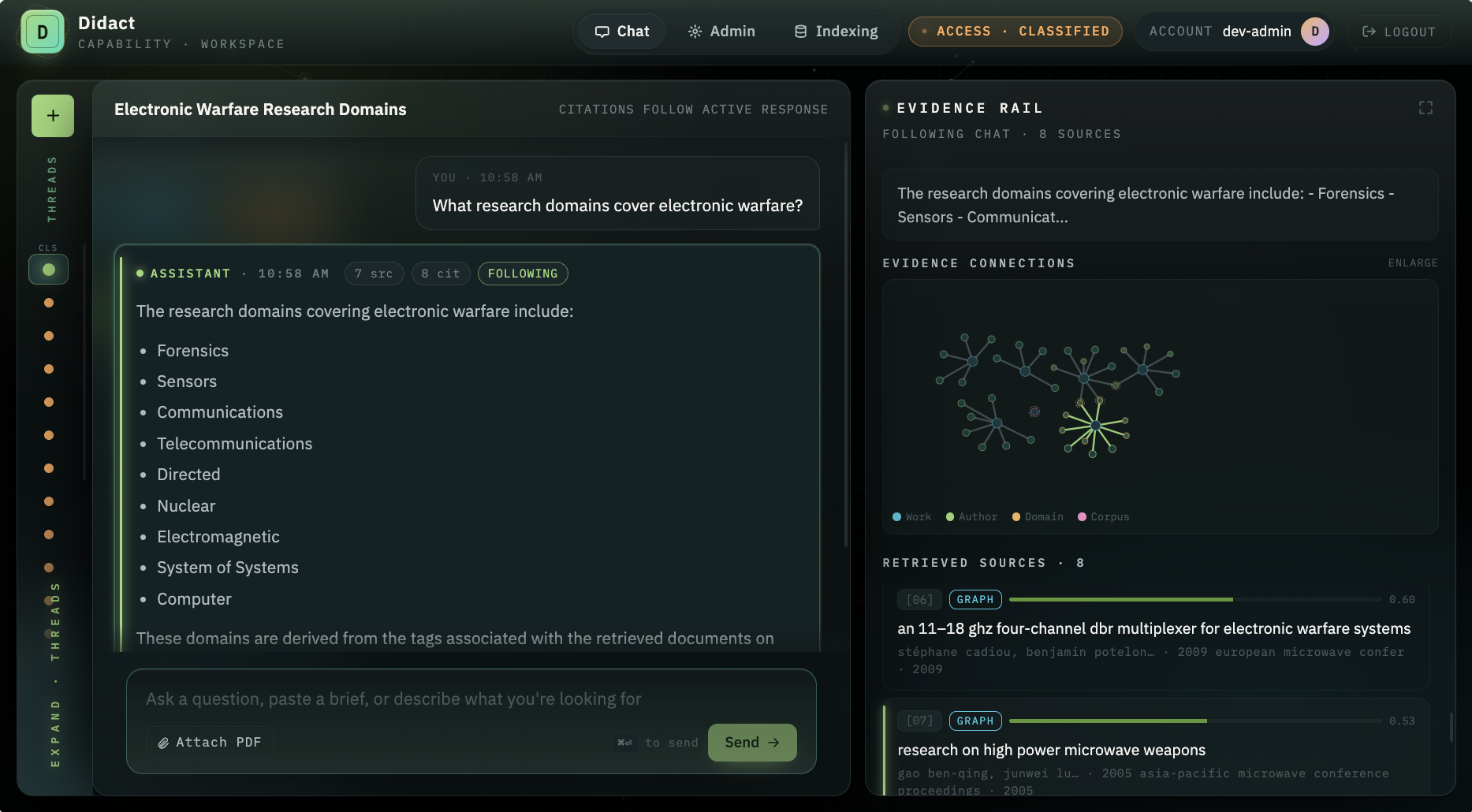}\hspace{0.05em}
        \includegraphics[width=0.4\linewidth]{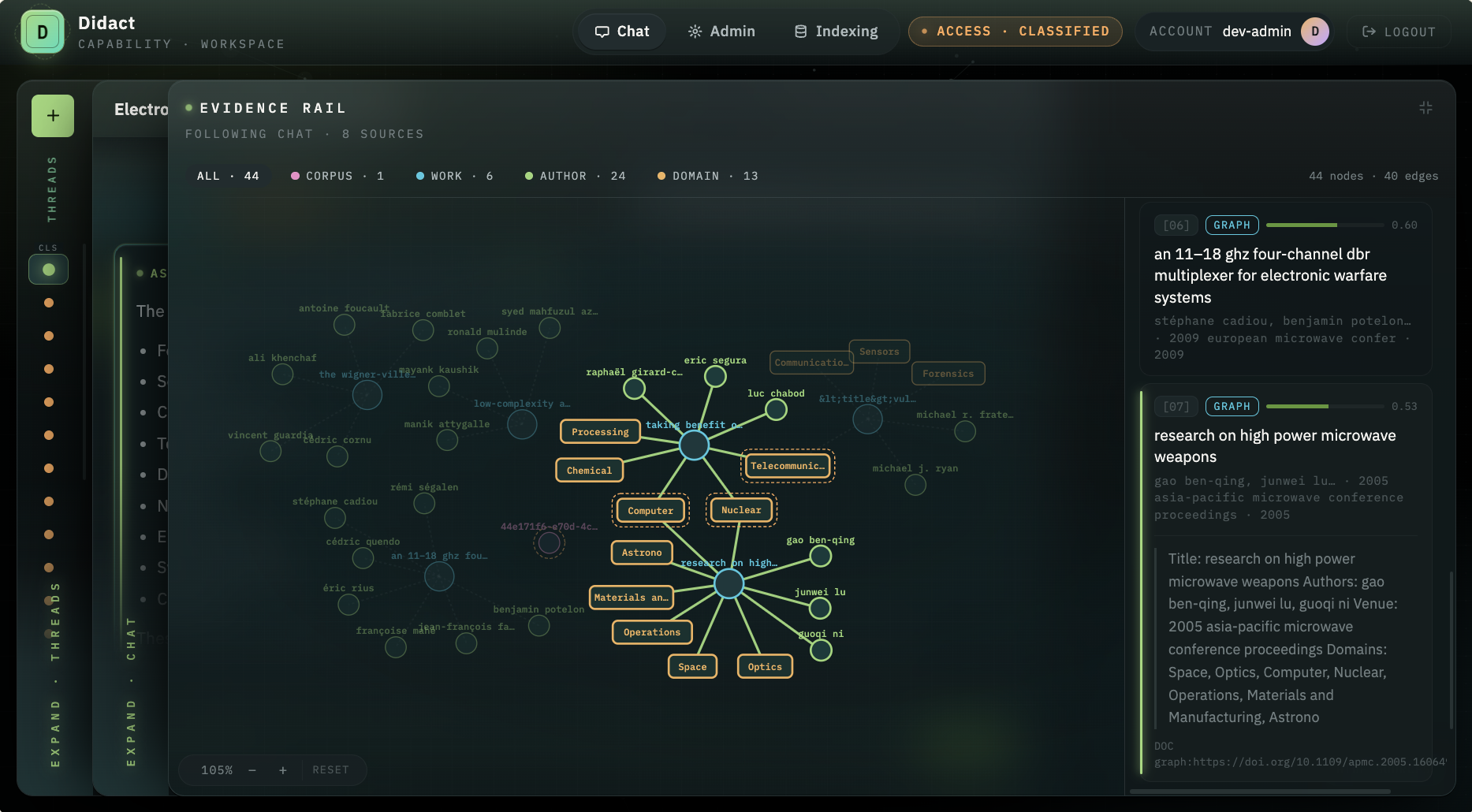}
  %  \fbox{\parbox{0.92\linewidth}{\centering \includegraphics[\linewidth]{didact-demo-1.png}}}
    \caption{Screenshots of Didact.}
   % \Description{Placeholder for the Didact system architecture figure.}
    \label{fig:screenshot}
\end{figure*}

\subsection{Evidence Rail Assembler}
The Evidence Rail is a key interface contribution of Didact. For each assistant response, it shows the focused message, source cards for retrieved document chunks and graph nodes, and a graph preview when graph retrieval was used. Expanding the rail reveals a dynamically generated graph over all retrieved sources, implemented with Cytoscape.js \cite{Franz_2015}. The graph layout is computed once per citation set and cached, so expanding the rail does not shuffle node positions. Users can zoom, pan, reposition individual nodes, and highlight one or more nodes; these interactions expand the corresponding source cards and scroll to them. The Evidence Rail, thus, transforms citations from a static list into an inspectable evidence workspace.
\subsection{Other Features}
In addition to the core AI architecture, Didact also provides the following features:
\begin{itemize}[leftmargin=*]
    \item \textbf{Persistence Layer}
A persistence layer saves each query turn together with its citations and retrieval results into a database. Storing citations separately from the generated text allows the frontend to reconstruct the Evidence Rail for any past turn and keeps source inspection available after a conversation is resumed.
\item \textbf{Admin Portal}
The admin route lets authorised users configure the system prompt and model-provider backend, while the indexing route supports document-store reindexing. These pages are visible only to admin-class accounts, separating end-user capability discovery from operational controls and allowing system maintainers to tune prompts, providers, and corpora during prototype evaluation.
\item \textbf{Authentication}
Authentication is enforced at the application boundary and requires an access key. The key is stored in browser session storage for the duration of the tab session, with the authenticated context persisting across page refreshes. The key also determines access level (public or classified), achieving source isolation over the authorised vector store before retrieval.
\end{itemize}

\subsection{Front-end}
Didact is delivered as a browser-based prototype (screenshots in Figure~\ref{fig:screenshot}) built as a React and TypeScript single-page application, bundled with Vite and styled with TailwindCSS. Client-side routing is handled by React Router, and server state is managed through TanStack Query for caching, stale-time control, and mutation handling. FastAPI~\cite{fastapi2018} is used for calls to Composite RAG. A reverse proxy handles production traffic routing and static asset delivery, presenting the system as a unified single-origin deployment.  The UI itself consists of a main `Chat' dashboard which has a panel on the left to select past conversations (`Threads'), conversations in the middle and the evidence rail to the right. The Threads panel supports conversation creation, renaming, title previewing, and deletion. The `Conversations' panel streams assistant responses, automatically updating the `Evidence Rail' panel on the right so the evidence panel always reflects the citations for the answer the user is reading. The evidence panel itself also allows for highly interactive graphs.
\begin{figure}[h!]
    \centering
    \includegraphics[width=0.8\linewidth]{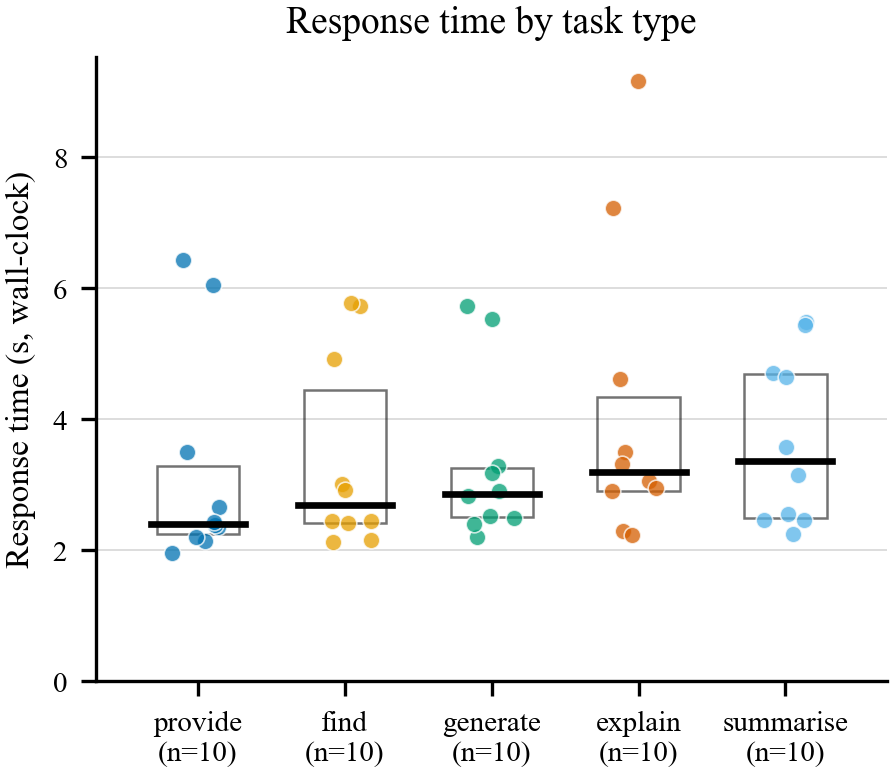}
  %  \fbox{\parbox{0.92\linewidth}{\centering \includegraphics[\linewidth]{didact-demo-1.png}}}
    \caption{Didact Response time by task type.}
   % \Description{Placeholder for the Didact system architecture figure.}
    \label{fig:res1}
\end{figure}

\begin{figure}[h!]
    \centering
    \includegraphics[width=0.75\linewidth]{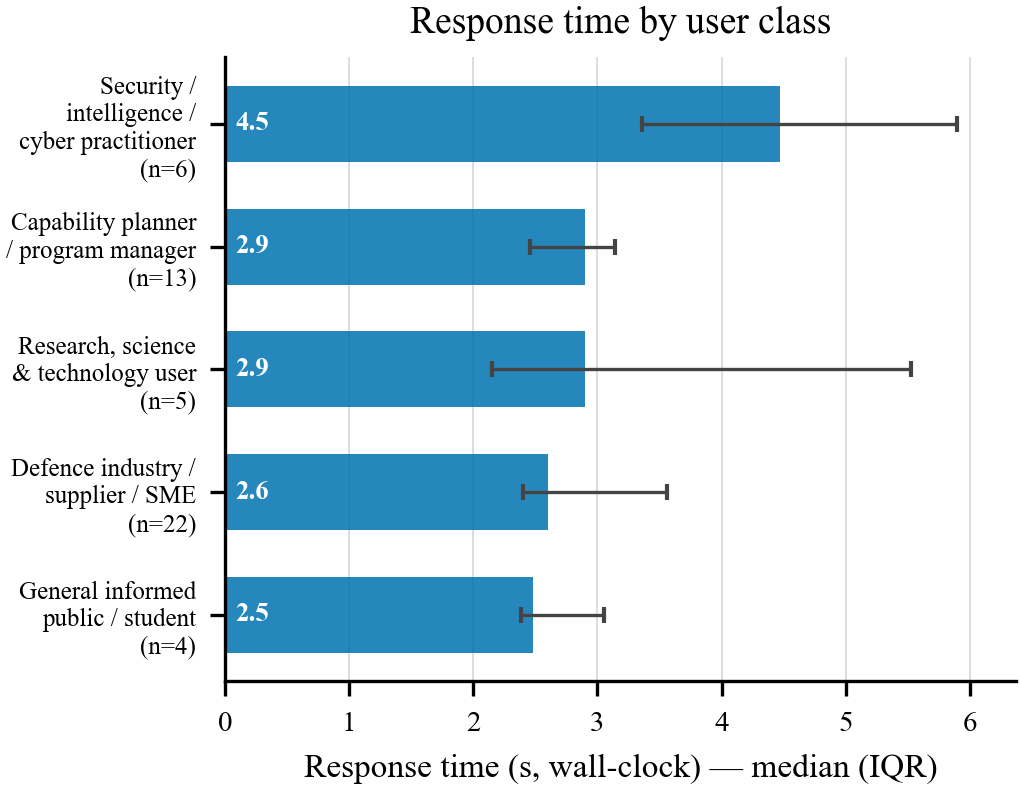}
  %  \fbox{\parbox{0.92\linewidth}{\centering \includegraphics[\linewidth]{didact-demo-1.png}}}
    \caption{Didact Response time by user persona.}
   % \Description{Placeholder for the Didact system architecture figure.}
    \label{fig:res2}
\end{figure}

\section{Evaluation}
\begin{table*}[h!]
\centering
\begin{adjustbox}{width=0.7\textwidth}
\begin{tabular}{lc}
\toprule
Metric & Score \\ \midrule
Hit@1 & 0.680 \\
Hit@2 & 0.760 \\
\bottomrule
\end{tabular} \hspace{2em}
\begin{tabular}{lcc}
\toprule
Metric & Mean & Std \\ \midrule
Context Recall    & 0.904 & 0.238 \\
Answer Relevancy  & 0.882 & 0.239 \\
Context Precision & 0.780 & 0.174 \\
\bottomrule
\end{tabular}\hspace{2em}
\begin{tabular}{lccc}
\toprule
Task & Hit@1 & Ctx.\ Recall & Ans.\ Relev. \\ \midrule
explain   & 0.600 & 1.000 & 0.820 \\
find      & 0.700 & 0.800 & 0.980 \\
generate  & 0.600 & 0.860 & 0.900 \\
provide   & 0.600 & 0.930 & 0.810 \\
summarise & 0.900 & 0.930 & 0.900 \\ \midrule
Overall   & 0.680 & 0.904 & 0.882 \\
\bottomrule
\end{tabular}
\end{adjustbox}
\caption{(a) Document retrieval accuracy against DoRA-QnA50~\cite{doan2026domain} ($n=50$); (b) RAGAS evaluation against DoRA-QnA50~\cite{doan2026domain} ($n=50$). GPT-4o-mini used as judge; (c) Per-task breakdown of key evaluation metrics ($n=10$ per task).}
\label{tab:ragas}
\end{table*}
\textbf{Run-time benchmarking}: Based on Didact's potential stakeholders in the defence ecosystem, we identified relevant user personas and task types drawn from~\cite{doan2026domain}. The wall-clock response time in seconds for five task types (provide, find, generate, explain, summarise) and five user personas are shown in Figs~\ref{fig:res1} and ~\ref{fig:res2} respectively. The reasonable response times across diverse usage scenarios suggest Didact's usability for a wide range of stakeholders.
% Each DoRA gold answer was independently scored by four domain experts across seven quality dimensions (comprehensiveness, relevance, groundedness, verbosity, composition, clarity, alignment), achieving an overall acceptance rate of 98\% and mean scores of 4.50--4.97 out of 5.
\textbf{Response quality benchmarking}: We also evaluate Didact using DoRA-QnA50 from DoRA~\cite{doan2026domain}, a domain-expert-validated benchmark of 50 question--answer pairs spanning the same five task types, five user personas and seven defense domains. Didact responses are evaluated against the DoRA gold answers and their associated retrieved contexts using document retrieval metrics and RAGAS~\cite{es2024ragas} LLM-as-judge metrics; results are shown in Tables~\ref{tab:ragas}. Didact achieves strong retrieval quality: Hit@1 of 0.680 shows the correct source document is retrieved as the top citation in 68\% of cases (Hit@2: 0.760), with \textit{summarise} tasks reaching Hit@1 = 0.900. The 16 misses largely involve thematically adjacent documents within the same series, suggesting topical confusion rather than retrieval failure. Context Recall of 0.904 confirms the retrieved content covers the gold answer in 90\% of cases regardless of exact document match, and Context Precision of 0.780 confirms retrieved passages are focused on the question. Answer Relevancy of 0.882 shows that responses directly address the user question. Across all 50 queries, Didact produces zero fallbacks, with a mean LLM response latency of 2.17\,s (95th percentile: 2.91\,s).

\textbf{Demonstration Scenario}: The demonstration will interest people from any professional background which requires exploration of multi-domain data, and will be particularly relevant for specialist-domain (such as Defence) stakeholders. An example user is a Defence capability manager who is responsible for assessing Australian research strength in an emerging national security priority area. They need to understand which organisations, researchers, and technology domains are relevant, and whether the evidence suggests potential collaboration pathways or capability gaps. These users operate in an environment where information is fragmented across Defence datasets and research information, and need rapid, evidence-based insight. Evidence rail and auditability in Didact are vital to ensure that findings can be explained to senior decision-makers, program owners and procurement stakeholders. Using Didact, the capability manager submits a question: ``What are the national security priority areas this year?" and then decides to explore one of them further by asking, ``Which Australian researchers and organisations are contributing to advanced sensing for contested environments, and what evidence shows potential Defence relevance or collaboration opportunities?" Didact retrieves relevant passages from Defence and policy documents, surfaces connected entities from the academic research graph, and presents these through the Evidence Rail.
\section{Conclusion}
Didact is a natural language conversation tool that uses composite RAG to support defence capability discovery across fragmented document and research sources. By combining access-controlled document retrieval, structured knowledge-graph retrieval, agent orchestration, and an interactive Evidence Rail, Didact's evidence rail feature preserves access boundaries and provides interactive source traceability, strengthening its reliability for the Defence setting. Didact demonstrates strong capability in terms of both the quality of output and running times validated over multiple task types and defence user personas.

\begin{acks}
Didact has been developed by University of New South Wales Sydney and Cyndr.ai, via a Defence Trailblazer grant awarded by Australian Government's Department of Education. 
\end{acks}

\section*{GenAI Usage Disclosure}
Generative AI was used only for minor grammar and spelling correction and stylistic suggestions during the creation of this manuscript and for bug-fixing and code refinement during the development.

\bibliographystyle{ACM-Reference-Format}
\bibliography{main}

\end{document}